\documentclass[letterpaper,10pt,conference]{ieeeconf}

\IEEEoverridecommandlockouts
\usepackage[utf8]{inputenc}
\usepackage[T1]{fontenc}
\usepackage{graphicx}
\usepackage{booktabs}
\usepackage{amsfonts}
\usepackage{amsmath}
\usepackage{amssymb}
\usepackage{algorithm}
\usepackage{algorithmic}
\usepackage{multirow}
\usepackage{placeins}
\usepackage{url}
\usepackage{xcolor}

\usepackage{cite}
\usepackage[hidelinks]{hyperref}
\usepackage[capitalise,noabbrev]{cleveref}
\crefname{figure}{Fig.}{Figs.}
\Crefname{figure}{Fig.}{Figs.}
\crefname{table}{Table}{Tables}
\Crefname{table}{Table}{Tables}
\crefname{section}{Sec.}{Secs.}
\Crefname{section}{Sec.}{Secs.}
\crefname{equation}{Eq.}{Eqs.}
\Crefname{equation}{Eq.}{Eqs.}

\usepackage{array}
\newcolumntype{L}[1]{>{\raggedright\arraybackslash}p{#1}}
\newcolumntype{C}[1]{>{\centering\arraybackslash}p{#1}}
\newcommand{\paperTitle}{Look Before You Act: Bridging Active Perception and Vision-Language-Action Models via Observation-Aligned Data Adaptation}
 \title{\LARGE \bf \paperTitle}
\author{Jiawei Geng$^{1}$, Zhenyang Liu$^{1,2}$, Yuzheng Wu$^{1}$, Yongchong Gu$^{1}$, Yanwei Fu$^{1,2}$\\[0.4em]
\normalsize $^{1}$Fudan University\qquad $^{2}$Shanghai Innovation Institute}
\hypersetup{
  pdftitle={\paperTitle},
  pdfauthor={Jiawei Geng, Zhenyang Liu, Yuzheng Wu, Yongchong Gu, Yanwei Fu},
  pdfsubject={Active perception and vision-language-action policy adaptation},
  pdfkeywords={robot manipulation, active perception, vision-language-action models, demonstration adaptation}
}
\begin{document}
\maketitle
\thispagestyle{plain}
\pagestyle{plain}
\begin{abstract}

Robots manipulating objects in clutter must decide not only how to act, but also where to look. Although active camera motion can expose an occluded target, it may move the robot into an initial observation–state distribution that differs from the manipulation policy’s training distribution. We identify this problem as \emph{first-frame distribution sensitivity}: under perturbed starts in LIBERO, naively prepending camera repositioning to $\pi_{0.5}$ reduces success from 46.3\% to 11.5\%. We introduce \textbf{ActiveAlign}, a data-adaptation framework that trains a Vision-Language-Action (VLA) policy from the states at which active perception terminates. A learned viewpoint module first generates a camera trajectory. Offline, ActiveAlign uses closed-loop alignment to connect the trajectory endpoint to an existing demonstration, replays the remaining manipulation trajectory, and retains successful continuations for fine-tuning. At deployment, the same viewpoint module precedes the adapted VLA without modifying its architecture. Across 40 LIBERO tasks, ActiveAlign achieves 70.7\% success with $\pi_0$ and 72.7\% with $\pi_{0.5}$, outperforming their standard-training baselines by 40.0\% and 26.4\% respectively. Matched-state and data-matched controls support gains beyond prefix shortening and generic viewpoint augmentation; OpenVLA and GR00T N1 also show the same trend. Success rate under physical occlusion in real world increases from 35/90 to 83/90. These results show that active perception benefits manipulation when policy training is aligned with the observation and robot-state distribution induced by viewpoint selection.

\end{abstract}

\begin{figure*}[!t]
\centering
\includegraphics[width=0.98\textwidth]{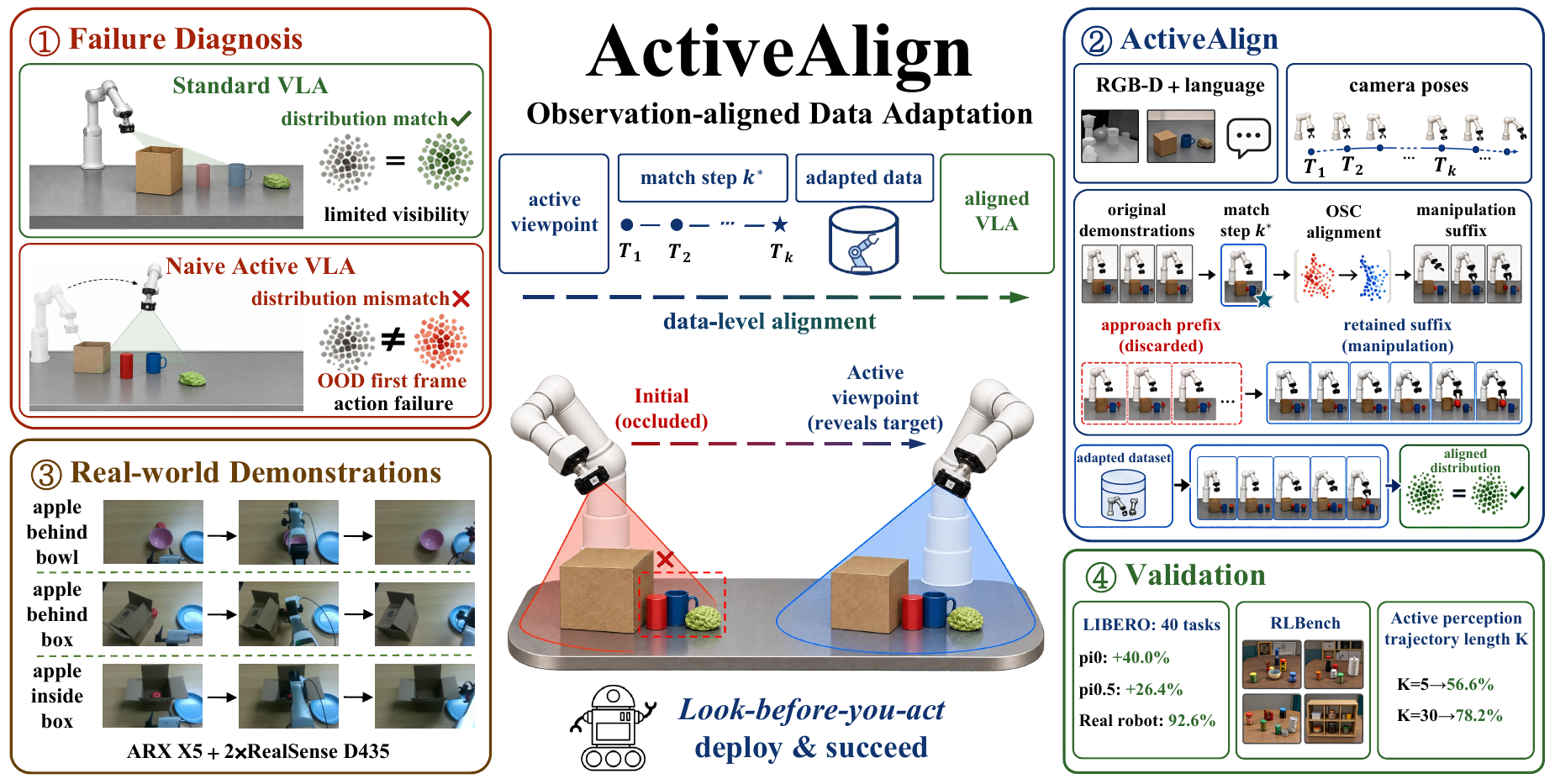}
\caption{\textbf{ActiveAlign overview.} Wrist-camera repositioning changes the view and robot state, potentially degrading an unadapted VLA. ActiveAlign fine-tunes the policy on successful trajectories beginning at active endpoints. The distribution sketches illustrate the intended alignment.}
\label{fig:teaser}
\end{figure*}

\section{INTRODUCTION}
Effective manipulation begins with an observation that supports the intended action~\cite{liu2025reasongrounder,liu2025neural}. A robot asked to retrieve fruit from a container or grasp a cup behind clutter may need to move its camera before it can see a useful approach direction. Vision-Language-Action (VLA) models~\cite{black2024pi0,kim2024openvla,intelligence2025pi05,team2024octo,liu2025trivla} provide a powerful interface from visual observations and language to robot actions, but their input is still determined by where the robot looks. Repositioning an existing wrist camera is an appealing way to expose the target before manipulation. The resulting system must therefore solve two connected problems: selecting an informative view and continuing the task from the state produced by that selection.

The most direct composition is to let an active-perception (AP) module improve the view, then hand control to a VLA trained on conventional demonstrations. Our experiments reveal a counterintuitive outcome: this composition can substantially reduce task success. Under a shared perturbed-start LIBERO protocol, $\pi_{0.5}$ falls from 46.3\% to 11.5\% average success after camera repositioning, and $\pi_0$ falls from 30.7\% to 7.0\%. The qualitative examples in \cref{fig:teaser} illustrate the tension: a policy can act from a familiar but occluded view, yet struggle to exploit a view selected to reveal the target. Improving the observation and making it usable by the action policy are distinct requirements.

We study this interface through \emph{first-frame distribution sensitivity}. Here, the first frame is the observation at the handoff from active perception to manipulation. Wrist-camera motion changes image location, scale, and occlusion together with arm configuration and the relationship to the remaining task. A policy must consequently infer a suitable continuation from a different combination of visual and proprioceptive cues. Action chunking~\cite{zhao2023aloha,chi2023diffusion} can propagate an early error through the actions executed before the next observation, even when later decisions use fresh images. This motivates examining the joint observation--action interface; a difference between first-frame image distributions alone does not establish the cause of every failure. Our experiments therefore pair the failure observation with controls that separate viewpoint information from robot-state progress and data volume.

The central design insight is that the deployment viewpoint policy is available during data construction. Rather than requiring a manipulation learner to accommodate arbitrary sensing motions, we can expose it to successful continuations from the endpoints of the very module it will encounter at inference. This turns the composition problem into a question about supervision: \emph{how can existing demonstrations be made to begin where active perception stops?} Demonstration-generation methods such as MimicGen and DemoGen~\cite{mandlekar2023mimicgen,xue2025demogen} motivate reusing recorded behavior, while the present setting adds a specific constraint: the starting observation and state are chosen by a task-conditioned camera policy. Visual augmentation alone does not supply the actions needed to continue from that state.

\textbf{ActiveAlign} constructs these continuations offline. Given scene RGBD and a language instruction, a learned module predicts a wrist-camera trajectory. In a restored environment, we execute the trajectory and match its endpoint to a nearby end-effector state in an early demonstration prefix. Closed-loop alignment bridges the residual gap, after which the recorded manipulation suffix is replayed. Successful alignment-and-replay trajectories become training data for the VLA. The active approach replaces part of the original demonstration, while the recorded continuation teaches the policy how to proceed after that approach. At deployment, the same viewpoint module runs once before object contact; the adapted VLA then manipulates using fresh observations. The procedure changes training data and policy weights while preserving the VLA architecture.

Replay feasibility is essential to this construction. End-effector proximity does not ensure that object and contact states agree, so nearest-state matching alone cannot validate a continuation. Successful replay filters the candidate endpoints and limits the training distribution to states for which usable supervision can be reconstructed. ActiveAlign consequently applies to resettable simulators and reconstructable physical workcells. This requirement belongs to offline data collection: deployment requires neither scene reset nor expert replay. The resulting scope is pre-manipulation view selection, with longer-horizon tasks testing the lasting value of that initial view rather than continuous response to moving occluders.

Our evaluation follows this reasoning from composition failure to controlled explanation and physical validation. On LIBERO Spatial, a matched-state camera-extrinsics control holds the demonstration step and manipulation suffix fixed and reaches 54.6\%, compared with 78.2\% for ActiveAlign. Data-matched random views and broad viewpoint jitter reach 58.0\% and 61.4\%, respectively. These comparisons support a benefit from the informative, deployment-matched observations beyond shorter demonstrations and generic coverage. Across four suites, ActiveAlign improves $\pi_0$ and $\pi_{0.5}$ by 40.0 and 26.4 percentage points; OpenVLA and GR00T N1 extend the evidence to different action formulations. Initial-occlusion simulation tasks, diverse physical tasks, and endpoint perturbations further characterize when the construction is useful and where it breaks down.

Our contributions are threefold:
\begin{itemize}
\item We characterize the failure of naive composition of active perception and VLAs, and test competing explanations through matched-state and data-matched controls.
\item We introduce observation-aligned demonstration adaptation, connecting active-view endpoints to successful manipulation through prefix matching, closed-loop alignment, and replay filtering.
\item We evaluate the construction across four VLA backbones, 40 LIBERO tasks, initial occlusion, and physical manipulation, with sensitivity studies that quantify the feasibility--diversity trade-off and endpoint error tolerance.
\end{itemize}

\section{RELATED WORK}
\noindent\textbf{Vision-language-action policies.}
Large-scale language-conditioned robot policies such as RT-1 and RT-2 established a general interface from visual observations and language instructions to robot actions~\cite{brohan2022rt1,brohan2023rt2}. Recent VLA models instantiate this interface with different action representations: $\pi_0$ and $\pi_{0.5}$ use flow matching~\cite{black2024pi0,intelligence2025pi05}, OpenVLA predicts discrete action tokens~\cite{kim2024openvla}, and GR00T N1 uses a diffusion-transformer action module~\cite{bjorck2025groot}. Diffusion policies and alternative action tokenizations provide additional formulations for visuomotor control~\cite{chi2023diffusion,ze20243d,pertsch2025fast,liu2025spatial}. Despite these architectural differences, each policy depends on the observation--state distribution represented in its training demonstrations. ActiveAlign is orthogonal to these action parameterizations: it preserves the policy architecture while adapting the fine-tuning distribution to the states induced by a deployment-time viewpoint module which offers a more informative view.

\noindent\textbf{Active and multi-view perception.}
Active perception~\cite{bajcsy1988active,chen2011active} and next-best-view methods~\cite{connolly1985determination,doumanoglou2016recovering} select sensing actions that improve downstream decisions. Multi-view manipulation methods aggregate or render observations from multiple viewpoints to improve spatial reasoning~\cite{morrison2019multi,goyal2023rvt,goyal2024rvt2,liu2026activevla}, while SAM2Act, VISO-Grasp, and ViewActive explore related mechanisms for visual representation and viewpoint selection~\cite{wang2024sam2act,chen2025visoGrasp,viewactive2024}. Our approach, instead, physically repositions a wrist camera before invoking a separately adapted VLA, thus studying physical wrist-camera motion, which changes not only the visual input but also the robot configuration from which manipulation begins. Its focus is therefore the compatibility between a selected better viewpoint and the downstream policy's manipulation distribution.

Recent robotic systems learn physical sensing motion together with manipulation. Vision in Action couples sensing and bimanual behavior using a 6-DoF neck and VR teleoperation~\cite{yu2023visioninaction}; HoMMI combines egocentric human demonstrations, hand--eye policies, and whole-body control~\cite{xu2026hommi}; and ActiveGlasses uses a dedicated perception arm to coordinate camera and manipulation motion~\cite{zou2026activeglasses}. These systems collect or learn paired sensing and manipulation behavior for specialized active-vision embodiments. In contrast, ActiveAlign uses an existing wrist camera for a single pre-manipulation viewpoint trajectory and connects it to an independently trained VLA through offline demonstration adaptation, without altering the VLA architecture or requiring continuous gaze control. GenDoP~\cite{zhang2025gendop} studies related autoregressive RGBD-and-language camera trajectory generation~\cite{liu2026omnicam}; our contribution concerns the robot-data adaptation interface.

\noindent\textbf{Demonstration adaptation and distribution shift.}
DAgger~\cite{ross2011dagger} addresses policy-induced state mismatch through iterative collection. Domain randomization and augmentation broaden visual coverage~\cite{tobin2017domain,laskin2020reinforcement,hansen2021generalization}, including through generated images~\cite{yu2023scaling,chen2023genaug,black2023demogen}. ActiveAlign targets a different source of shift: the handoff states produced by a separate viewpoint policy used at deployment. Demonstration-generation methods reuse recorded behavior to expand the training distribution. MimicGen transforms demonstrations across object configurations~\cite{mandlekar2023mimicgen}, while DemoGen adapts trajectories and point-cloud observations for spatial generalization~\cite{xue2025demogen}. ActiveAlign instead adapts demonstrations to the physical endpoints induced by a task-conditioned viewpoint policy. Because camera motion changes both arm configuration and visual observations, geometric endpoint matching alone does not guarantee a valid manipulation continuation. ActiveAlign therefore uses closed-loop alignment and successful replay to validate supervision. Our matched-state and data-matched controls further compare this targeted, deployment-matched adaptation with shorter manipulation suffixes and broader random-view or viewpoint-jitter coverage.

\section{METHOD}
\label{sec:method}
\subsection{Problem Formulation}
\label{sec:problem}

Let $x_t$ denote the physical scene and robot state and let $o_t$ denote the complete policy observation, including scene RGB, wrist RGB, and proprioception where supported by the backbone. A policy $\pi_\theta$ conditions on $o_t$ and instruction $l$. Demonstrations contain state-observation-action trajectories; denote the observation distribution at original demonstration starts by $p_{\mathrm{demo},0}$.

An active perception module $\mathcal A$ generates a camera trajectory from a scene observation $s_0$ and instruction $l$. Execution induces a new state $\tilde x_0$ and observation $\tilde o_0$. The corresponding initial distribution $p_{\mathcal A,0}$ may differ from $p_{\mathrm{demo},0}$. Because the camera is wrist-mounted, this shift generally involves both vision and robot state. A mismatch of initial-image marginals alone neither proves that the input is outside the full training support nor establishes the cause of a policy failure.

Our objective is to improve manipulation from the states induced by $\mathcal A$. We construct adapted demonstrations $\widetilde{\mathcal D}$ and fine-tune with the backbone's supervised learning objective:
\begin{equation}
 \theta^*=\arg\min_\theta
 \mathbb E_{(o,l,a)\sim\mathcal D_{\mathrm{train}}}
 \bigl[\mathcal L_{\mathrm{VLA}}(\theta;o,l,a)\bigr].
 \label{eq:objective}
\end{equation}
Here $\mathcal D_{\mathrm{train}}$ denotes the condition's fine-tuning data, including aligned demonstrations for ActiveAlign; data-matched controls use the same mixing ratio. The target $a$ is an action or chunk, and $\mathcal L_{\mathrm{VLA}}$ is its token-prediction or continuous-action loss. Task success is evaluated through rollouts, not optimized directly by this loss.

The construction must preserve more than the appearance of the active view. Each observation needs an action target that is executable from the corresponding robot and scene state. We therefore use the deployed perception module to generate candidate starting states, and use recorded manipulation to supply their continuations. Prefix matching identifies a nearby continuation, alignment connects to it, and replay checks whether the resulting episode remains successful. These three operations jointly determine which candidate views can become useful VLA supervision.

\begin{figure*}[!t]
\centering
\includegraphics[width=0.98\textwidth]{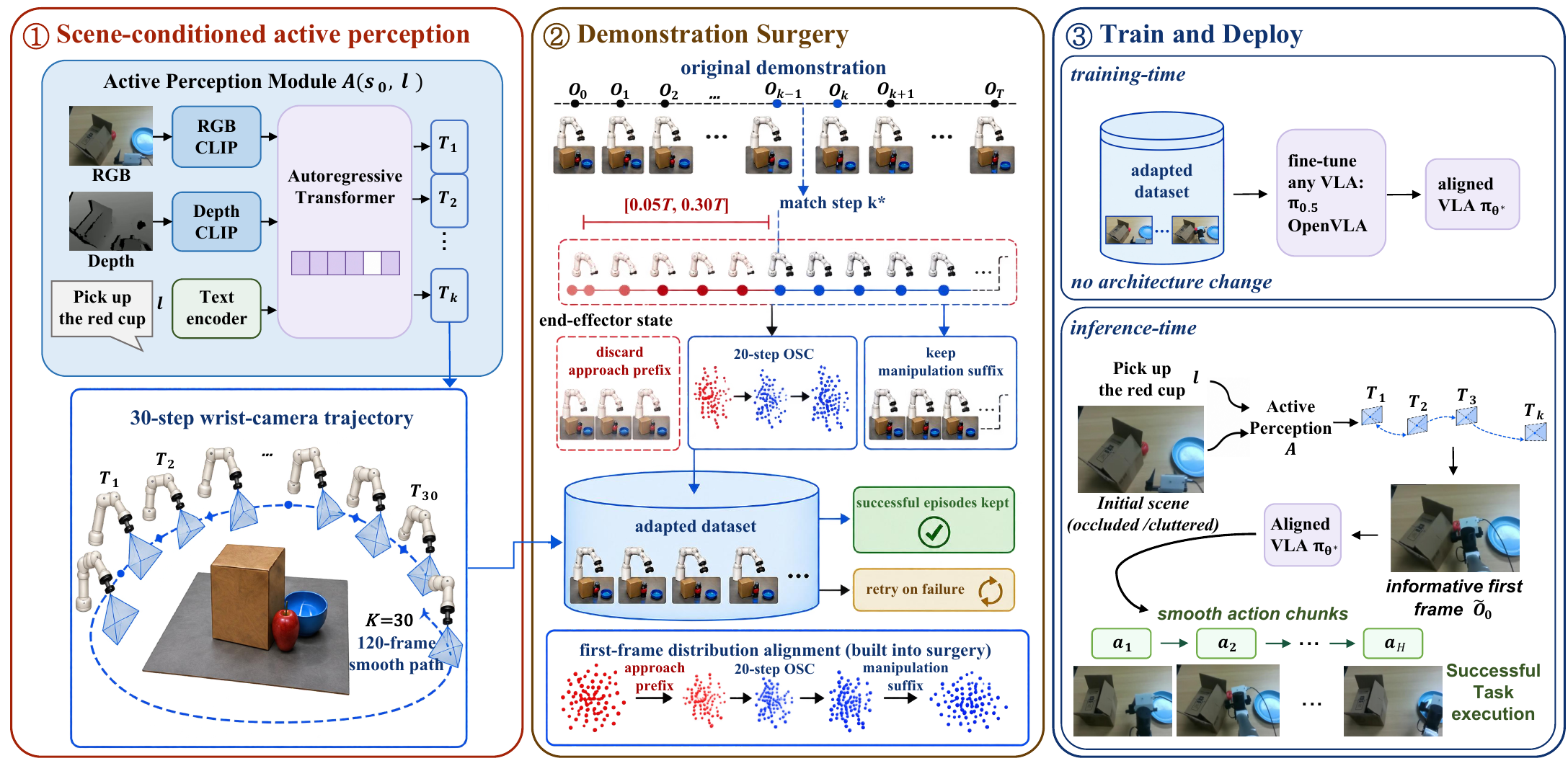}
\caption{\textbf{ActiveAlign pipeline.} Left: RGBD and language condition a camera trajectory. Center: prefix matching, alignment, and successful replay produce adapted demonstrations. Right: VLA fine-tuning and deployment, with one viewpoint-selection phase followed by manipulation from fresh observations. Physical motion is continuous; IK repositioning is simulation-only. Distribution sketches are schematic.}
\label{fig:pipeline}
\end{figure*}

\subsection{Active Perception Module}
\label{sec:active_perception}

View selection is conditioned on both the task and the scene: the instruction identifies what needs to be observed, while the surrounding geometry determines which camera motion may expose it. The module combines RGB features, depth features, and language features to predict a sequence of camera poses. RGB and depth use separate CLIP branches, and a conditional autoregressive decoder generates quantized pose tokens. The decoded extrinsics are passed to the robot controller, which turns the visual prediction into a candidate sensing motion.

Formally, the input is a scene RGBD observation $s_0=(I_{\mathrm{rgb}},D)$ and instruction $l$, and the output is $\{T_1,\ldots,T_K\}$. The default uses $K=30$ poses (\cref{fig:pipeline}), represented in $SE(3)$ and transformed into the calibrated robot/world frame. The generated sequence can be interpolated to 120 poses for smooth execution. Training uses scene-instruction-camera-trajectory pairs: 20,000 in simulation and 1,200 for the physical setting, separate from the VLA demonstrations.

Predicted paths are candidate sensing motions, without certified optimality or collision freedom. Simulation uses IK repositioning; the physical robot moves continuously through its controller. Adaptation retains endpoints permitting alignment and successful manipulation replay.

\subsection{Observation-Aligned Data Adaptation}
\label{sec:data_adapt}
In \cref{alg:adaptation}, the environment is restored to each demonstration's starting state. After executing $\mathcal A$, we identify the nearest end-effector state within a prefix window $[\alpha T,\beta T]$. Closed-loop Cartesian alignment closes the residual gap, after which the recorded manipulation suffix is replayed. Only the alignment segment and suffix of successful attempts are retained; the active approach is not a VLA training target.

\begin{algorithm}[t]
\caption{Observation-Aligned Data Adaptation}
\label{alg:adaptation}
\begin{algorithmic}[1]
\REQUIRE Demonstrations $\mathcal D$, viewpoint module $\mathcal A$, restorable environment $\mathcal E$, retry configurations $\mathcal C$
\STATE $\widetilde{\mathcal D}\leftarrow\emptyset$
\FOR{each demonstration $\tau^i\in\mathcal D$}
 \FOR{each configuration $c\in\mathcal C$}
  \STATE Restore $\mathcal E$ to the starting scene and robot state
  \STATE Capture $s_0$; predict camera trajectory using $\mathcal A(s_0,l^i)$
  \STATE Execute candidate trajectory under configuration $c$
  \STATE Record endpoint end-effector state $\tilde q_{\mathrm{ee}}$
  \STATE Find nearest demo state $k^*$ in $[\alpha T_i,\beta T_i]$
  \STATE Record closed-loop alignment and subsequent suffix replay
  \IF{the episode completes successfully}
   \STATE Add the recorded alignment and replay to $\widetilde{\mathcal D}$
   \STATE \textbf{break}
  \ENDIF
 \ENDFOR
\ENDFOR
\RETURN $\widetilde{\mathcal D}$
\end{algorithmic}
\end{algorithm}

\noindent\textbf{Prefix matching.}
The default window is $[0.05T,0.30T]$. Earlier matching retains more of the original approach; later matching shortens manipulation and may require unreproduced scene changes. End-effector proximity does not ensure matching object/contact states, making successful replay a necessary filter. Our window study measures retention, diversity, and success.

\noindent\textbf{Closed-loop alignment.}
A recorded adjustment connects the active endpoint to the matched suffix using operational space control (OSC) in simulation and Cartesian control on the robot. The default simulation adjustment budget is 20 steps. Directly replaying the suffix from the unmatched endpoint can accumulate control error, so this transition is included in the adapted trajectory. The VLA therefore receives supervision for leaving the active endpoint as well as following the retained manipulation. Unlike IK repositioning, the alignment segment consists of executable VLA training actions.

\noindent\textbf{Filtering and retries.}
Simulation retries vary the active trajectory budget and alignment steps. The reported adaptation counts are 1,736 retained out of 2,000 attempted in simulation (86.8\%) and 86 retained out of 100 attempted on the physical platform (86.0\%).

Consequently, $\widetilde{\mathcal D}$ represents a \emph{feasibility-filtered active-view distribution}: filtering and retries can make $p_{\widetilde{\mathcal D},0}$ differ from $p_{\mathcal A,0}$. Sharing the viewpoint module targets nominal alignment, without guaranteeing distribution equality or optimal task success. The control and perturbation studies test its practical benefit and sensitivity.

\subsection{Deployment and Applicability}
\label{sec:deployment}
Deployment consists of one active approach followed by manipulation. The viewpoint module is queried once before object contact. Thereafter, the VLA obtains fresh images and state inputs and predicts actions. For the chunked $\pi_0$ and $\pi_{0.5}$ implementations,
\begin{equation}
 \{a_t,\ldots,a_{t+H-1}\}=\pi_{\theta^*}(o_t,l),
 \quad H=10,\quad R=5,
\end{equation}
where the first $R$ actions are executed before replanning. Thus errors can accumulate within the executed portion, but the policy does not commit to all $H$ actions without correction. Other backbones retain their native action-generation interfaces.

The wrist camera moves with the arm during manipulation, but no further viewpoint-selection phase is invoked. Offline adaptation needs simulation reset or a reconstructable physical workcell; deployment needs neither reset nor demonstration replay. Replacing $\mathcal A$ may require regenerating data and fine-tuning.

\section{EXPERIMENTS}
\label{sec:experiments}
\subsection{Experimental Setup}
\noindent\textbf{Environments and data.}
LIBERO~\cite{liu2024libero} provides four suites of 10 tasks, with 50 original demonstrations per task (2,000 total). LIBERO-Occ~\cite{li2026liberoocc} uses the same suites but occludes the target in the default initial view. The physical evaluation runs on two arm settings (an ARX X5 arm and an ARCHER-Y6 arm), both with scene-level and wrist-mounted Intel RealSense D435 cameras, and 100 original demonstrations. RLBench~\cite{james2020rlbench} is used only to visualize zero-shot transfer of the perception module.

\noindent\textbf{Backbones and training.}
We evaluate $\pi_0$, $\pi_{0.5}$, OpenVLA, and GR00T N1. We follow the model-specific recommended training recipes, with hyperparameters selected using the validation set. All models share the LIBERO Franka end-effector delta-pose and gripper interface.

\noindent\textbf{Perturbed-start protocol.}
Each LIBERO result uses 50 rollouts per task, with horizons of 220, 280, 300, and 520 steps for Spatial, Object, Goal, and Long. Initial end-effector pose coordinates receive zero-mean Gaussian perturbations with standard deviation 0.1 (translation in meters and rotation in radians). This perturbed-start protocol differs from the benchmark's prescribed initial states and is held fixed across all method conditions.

\noindent\textbf{Conditions and statistical reporting.}
\emph{Standard} trains on original demonstrations and uses no AP; \emph{Naive Active} invokes AP only at inference; \emph{ActiveAlign} trains on observation-aligned adaptations and invokes the same AP module. Data-matched controls fix retained episode count, filtering, suffix length, optimization, and mixing. LIBERO uses the environment completion checker. Physical results report binary counts and Wilson 95\% confidence intervals.

\subsection{Simulation Results and Analysis}
\label{sec:exp_sim}

\noindent\textbf{Perturbed-start performance.} 
\Cref{tab:libero_main} shows the handoff failure directly: AP without aligned training reduces average success from 30.7\% to 7.0\% for $\pi_0$ and from 46.3\% to 11.5\% for $\pi_{0.5}$. ActiveAlign instead reaches 70.7\% and 72.7\%, gains of 40.0 and 26.4 percentage points (pp) over Standard, with improvements on every suite. These values are comparisons within our perturbed-start protocol.

\begin{table}[t]
\caption{Perturbed-start LIBERO success rate(\%). }
\label{tab:libero_main}
\centering\footnotesize
\setlength{\tabcolsep}{2.4pt}
\begin{tabular}{@{}llrrrrr@{}}
\toprule
VLA & Setting & Spatial & Object & Goal & Long & Avg.\\
\midrule
\multirow{3}{*}{$\pi_0$}
&Standard&34.0&32.2&28.4&28.0&30.7\\
&Naive Active&6.6&0.0&13.6&7.6&7.0\\
&\textbf{ActiveAlign}&\textbf{73.8}&\textbf{78.2}&\textbf{66.6}&\textbf{64.2}&\textbf{70.7}\\
\midrule
\multirow{3}{*}{$\pi_{0.5}$}
&Standard&48.0&39.0&49.4&48.8&46.3\\
&Naive Active&22.0&2.2&18.6&3.0&11.5\\
&\textbf{ActiveAlign}&\textbf{78.2}&\textbf{78.0}&\textbf{68.8}&\textbf{65.8}&\textbf{72.7}\\
\bottomrule
\end{tabular}
\end{table}

The gains also extend beyond recovering the loss from naive composition. On Object, for example, ActiveAlign exceeds Standard by 46.0 pp for $\pi_0$ and 39.0 pp for $\pi_{0.5}$, while Naive Active reaches only 0.0\% and 2.2\%. Thus, the informative starting condition becomes useful when paired with adapted supervision. The two adapted backbones finish within 2.0 pp of each other on average; this motivates studying the data interface alongside model choice, without inferring that model capacity is generally irrelevant.

\begin{table}[t]
\caption{Backbone transfer on LIBERO (suite-average success rate, \%).}
\label{tab:backbones}
\centering\small
\begin{tabular}{lrrr}
\toprule
Backbone & Standard & Naive Active & ActiveAlign\\
\midrule
$\pi_0$ &30.7&7.0&\textbf{70.7}\\
$\pi_{0.5}$ &46.3&11.5&\textbf{72.7}\\
OpenVLA &56.4&24.8&\textbf{69.6}\\
GR00T N1 &51.8&21.6&\textbf{67.8}\\
\bottomrule
\end{tabular}
\end{table}

\Cref{tab:backbones} tests whether the adaptation is tied to the two flow-matching $\pi$ policies. ActiveAlign improves OpenVLA by 13.2 pp and GR00T N1 by 16.0 pp over their Standard conditions. The consistent trend across flow-matching, autoregressive-token, and diffusion-transformer policies supports compatibility with the tested action formulations, without implying universal backbone independence.

\Cref{tab:start_perturb} shows that under the benchmark-prescribed initial states, Standard and ActiveAlign are nearly matched; as the start-state shift increases, the gap widens.

\begin{table}[t]
\caption{Start-state shift sweep on LIBERO with $\pi_{0.5}$ (suite-average success rate, \%).}
\label{tab:start_perturb}
\centering\small
\setlength{\tabcolsep}{3pt}
\begin{tabular}{@{}lrrrr@{}}
\toprule
Start noise $\sigma$ (m/rad) & Benchmark init & 0.02 & 0.05 & 0.10\\
\midrule
Standard & 96.8 & 85.2 & 70.4&46.3\\
Naive Active & 48.0 & 38.2 & 23.1 & 11.5\\
\textbf{ActiveAlign} & 96.6 & 90.4 & 83.0 & 72.7\\
\midrule
\textbf{ActiveAlign} $-$ Standard& $-$0.2 & +5.2 & +12.6 & +26.4 \\

\bottomrule
\end{tabular}

\end{table}

\noindent\textbf{Viewpoint and data controls.} \Cref{tab:controls} isolates two competing explanations. The matched-state control fixes $k^*$, suffix length, retained count, training schedule, and mixing, while independently changing camera extrinsics to provide a default view instead of an active informative observation. Its 54.6\% success, versus 78.2\% for ActiveAlign, leaves a 23.6 pp gap after controlling for prefix shortening. As an auxiliary diagnostic, the target is visible in 43.7\% of the matched-state control observations and 74.8\% of the corresponding ActiveAlign observations.This visibility difference confirms that the two conditions provide substantially different target information; together with the controlled success comparison, it is consistent with the benefit of training on informative views induced by the deployment viewpoint module. Random-view adaptation and broad viewpoint jitter adaptation use the same data and optimization budgets and reach 58.0\% and 61.4\%. Generic viewpoint coverage therefore improves over Standard, but remains 20.2 pp and 16.8 pp below ActiveAlign. These results support the value of deployment-matched viewpoint adaptation beyond additional data, shorter continuations, and untargeted viewpoint diversity.

\begin{table}[t]
\caption{Viewpoint and data controls on LIBERO Spatial with $\pi_{0.5}$.}
\label{tab:controls}
\centering\small
\begin{tabular}{lr}
\toprule
Variant & Success (\%) \\
\midrule
Standard &48.0\\
Naive Active &22.0\\
Matched-state control &54.6\\
Random-view adaptation &58.0\\
Broad viewpoint jitter &61.4\\
\textbf{ActiveAlign} &\textbf{78.2}\\
\bottomrule
\end{tabular}
\end{table}

\noindent\textbf{Initial occlusion and perception transfer.} \Cref{tab:occ} evaluates targets hidden in the default first frame. ActiveAlign exceeds Standard by 28.6, 24.6, 22.4, and 34.6 pp on Spatial, Object, Goal, and Long. Naive Active remains below Standard. The Long result shows that a better initial observation can remain useful over a longer manipulation horizon; it does not test occluders that move during execution.

\begin{table}[t]
\caption{Initial-occlusion success (\%) on LIBERO-Occ with $\pi_{0.5}$. }
\label{tab:occ}
\centering\small
\begin{tabular}{lrrrr}
\toprule
Method & Spatial & Object & Goal & Long\\
\midrule
Standard &20.0&22.6&21.4&7.8\\
Naive Active &5.8&2.2&8.6&0.2\\
\textbf{ActiveAlign} &\textbf{48.6}&\textbf{47.2}&\textbf{43.8}&\textbf{42.4}\\
\bottomrule
\end{tabular}
\end{table}

\begin{figure}[t]
\centering
\includegraphics[width=\linewidth]{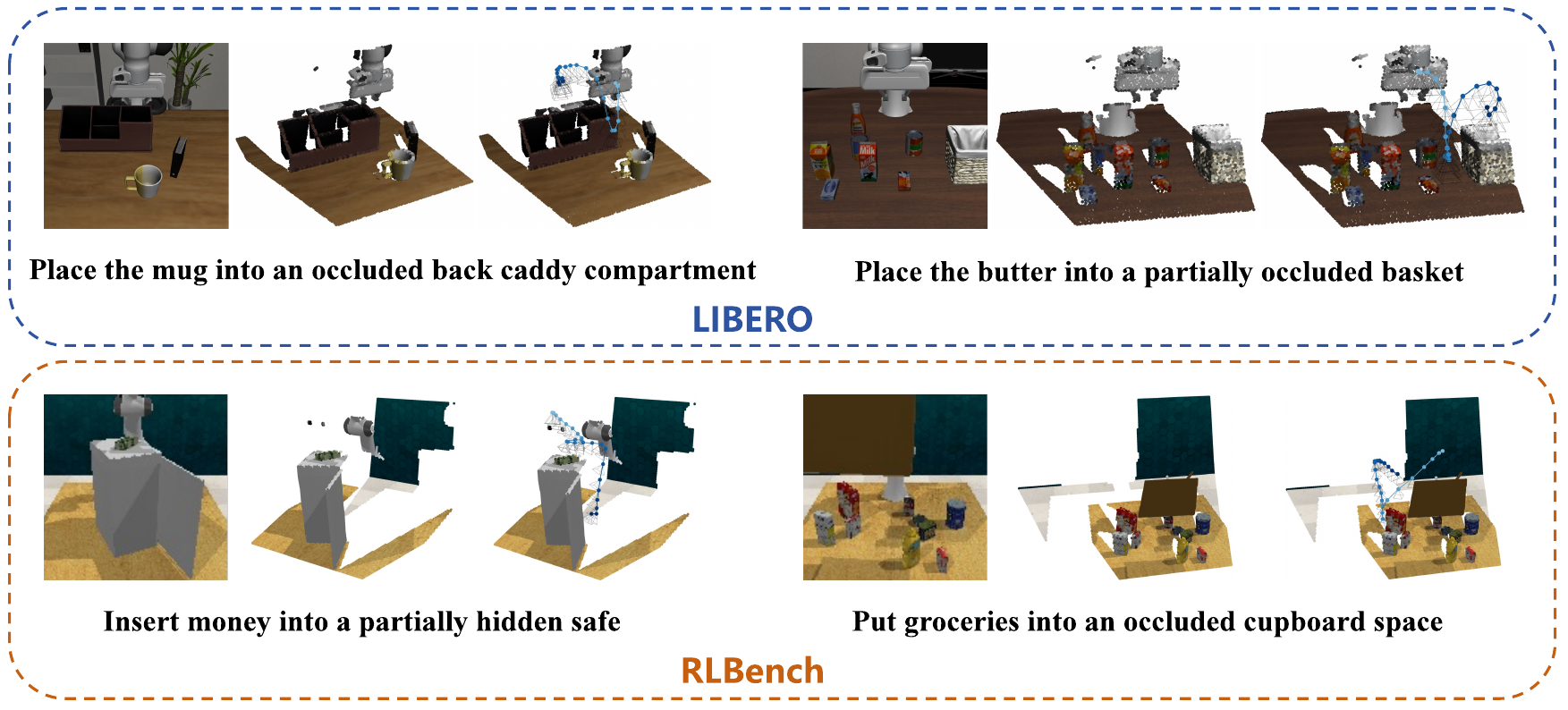}
\caption{\textbf{Perception trajectories.} Top: LIBERO. Bottom: perception-only zero-shot RLBench transfer. The examples illustrate view selection; end-to-end success is evaluated separately.}
\label{fig:ap_qual}
\end{figure}

\Cref{fig:ap_qual} shows the selected views around clutter. RLBench is a qualitative test of perception transfer only; end-to-end manipulation is evaluated in LIBERO and on the physical robot.

\begin{figure}[!t]
\centering
\includegraphics[width=\linewidth]{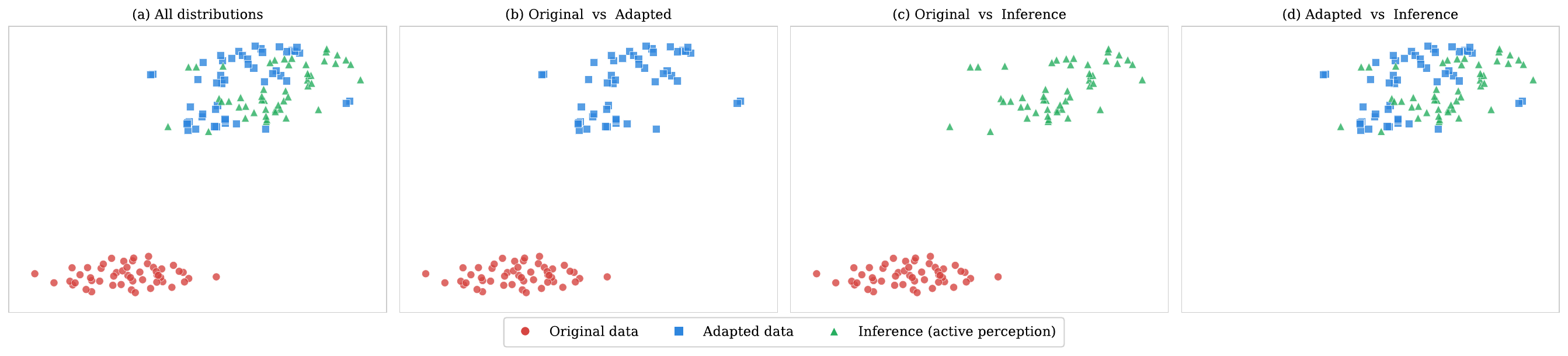}
\caption{\textbf{First-frame features.} Shared t-SNE embedding of CLIP ViT-H/14 features: red circles, original data; blue squares, adapted data; green triangles, inference after AP. Feature overlap is a qualitative illustration.}
\label{fig:tsne}
\end{figure}

In \cref{fig:tsne}, adapted demonstrations and AP-conditioned inference occupy overlapping feature regions. This visualization is descriptive; the controlled rollouts provide the causal evidence.

\begin{table*}[t]
\caption{Sensitivity on LIBERO Spatial with $\pi_{0.5}$. Each panel varies one factor at a time.}
\label{tab:sensitivity}
\centering\footnotesize

\begin{minipage}[t]{0.42\textwidth}
\centering
\textbf{(a) Offline matching window}\\[2pt]
\setlength{\tabcolsep}{3.0pt}
\begin{tabular}{@{}lrrr@{}}
\toprule
Window & Retained (\%) $\uparrow$ & Diversity $\uparrow$ & Success (\%) $\uparrow$\\
\midrule
$[0.00T,0.20T]$ & 81.4 & \textbf{0.116} & 75.8\\
$[0.05T,0.30T]$ & 86.8 & 0.108 & \textbf{78.2}\\
$[0.10T,0.40T]$ & \textbf{88.2} & 0.096 & 76.6\\
\bottomrule
\end{tabular}
\end{minipage}\hspace{0.5mm}
\begin{minipage}[t]{0.27\textwidth}
\centering
\textbf{(b) Terminal-position noise}\\[2pt]
\setlength{\tabcolsep}{5.0pt}
\begin{tabular}{@{}rr@{}}
\toprule
Noise $\sigma_p$ (m) & Success (\%) $\uparrow$\\
\midrule
0    & \textbf{78.2}\\
0.02 & 72.2\\
0.04 & 69.2\\
0.10 & 42.6\\
\bottomrule
\end{tabular}
\end{minipage}\hspace{0.5mm}
\begin{minipage}[t]{0.27\textwidth}
\centering
\textbf{(c) Viewpoint-trajectory length}\\[2pt]
\setlength{\tabcolsep}{7.0pt}
\begin{tabular}{@{}rr@{}}
\toprule
Generated poses $K$ & Success (\%) $\uparrow$\\
\midrule
5  & 56.6\\
10 & 61.2\\
20 & 71.0\\
30 & \textbf{78.2}\\
\bottomrule
\end{tabular}
\end{minipage}
\end{table*}

\begin{table*}[t]
\caption{Physical task diversity. Entries are successes/trials followed by Wilson 95\% confidence intervals in percentage points.}
\label{tab:real_diverse}
\centering
\setlength{\tabcolsep}{3pt}
\begin{tabular}{@{}L{0.25\textwidth}L{0.18\textwidth}C{0.15\textwidth}C{0.15\textwidth}C{0.15\textwidth}@{}}
\toprule
Task & Primary variation & Standard & Naive Active & ActiveAlign\\
\midrule
Grasp a cup on a cluttered tabletop &Target identity and clutter&9/20 [25.8,65.8]&4/20 [8.1,41.6]&\textbf{17/20 [64.0,94.8]}\\
Retrieve puzzles from a container &Object and container geometry&6/20 [14.5,51.9]&3/20 [5.2,36.0]&\textbf{15/20 [53.1,88.8]}\\
Observe, open, grasp, and place &Longer manipulation horizon&5/20 [11.2,46.9]&3/20 [5.2,36.0]&\textbf{13/20 [43.3,81.9]}\\
\bottomrule
\end{tabular}
\end{table*}

\noindent\textbf{Sensitivity analysis.} \Cref{tab:sensitivity} Panel (a) exposes the feasibility--diversity trade-off: later windows retain more adaptations but reduce endpoint diversity (measured as mean pairwise terminal-camera translation after scene normalization), and the middle window gives the highest success. Panel (b) perturbs only the predicted terminal camera position at inference. This perturbation is separate from the base protocol's initial robot-pose noise and tests sensitivity to errors in the AP endpoint. Success falls by 6.0 pp at 0.02 m and 9.0 pp at 0.04 m, then drops below Standard at 0.10 m. The method therefore tolerates moderate translational endpoint error but degrades when the deployment handoff moves substantially beyond the aligned training distribution. Panel (c) shows consistent gains as the trajectory budget grows from 5 to 30. This trend shows that performance benefits from additional viewpoint-trajectory capacity.

\subsection{Physical Evaluation}
\label{sec:exp_real}
\noindent\textbf{Collection and reset.}
Marked fixtures restore object and container states within 1 cm and $3^\circ$. Predicted camera paths are executed continuously; the recorded end-effector prefix is matched, closed-loop Cartesian control reduces the residual, and the manipulation suffix is replayed. An operator provides safety supervision only: near collision or force limits, the operator can abort but cannot correct a motion, teleoperate, or finish the task. The procedure retains 86 of 100 attempts. Reset and replay are used only for offline data construction.

\noindent\textbf{Evaluation protocols.}
We conduct two separate physical evaluations. The first evaluation measures task diversity using cup grasping, puzzle retrieval, and a longer observe--open--grasp--place task. Each task is evaluated for 20 trials, yielding 60 additional trials per method.  The second measures robustness to initial occlusion using three apple-manipulation tasks and three occlusion severities. Each task--severity combination is evaluated for 10 trials, yielding \(
  3\ \text{tasks}\times 3\ \text{severities}\times 10\ \text{trials}
  =90\ \text{trials per method}.
  \)

\begin{table}[t]
\caption{Physical occlusion success with $\pi_{0.5}$. Entries are successes/trials (rate in \%).}
\label{tab:real}
\centering\small
\setlength{\tabcolsep}{3pt}
\begin{tabular}{lccc}
\toprule
Occlusion & Standard & Naive Active & ActiveAlign\\
\midrule
Partial &16/30 (53.3)&8/30 (26.7)&\textbf{29/30 (96.7)}\\
Medium &11/30 (36.7)&5/30 (16.7)&\textbf{28/30 (93.3)}\\
Severe &8/30 (26.7)&4/30 (13.3)&\textbf{26/30 (86.7)}\\
\midrule
\textbf{Overall} &35/90 (38.9)&17/90 (18.9)&\textbf{83/90 (92.2)}\\
\bottomrule
\end{tabular}
\end{table}

\begin{figure}[t]
\centering
\includegraphics[width=\linewidth]{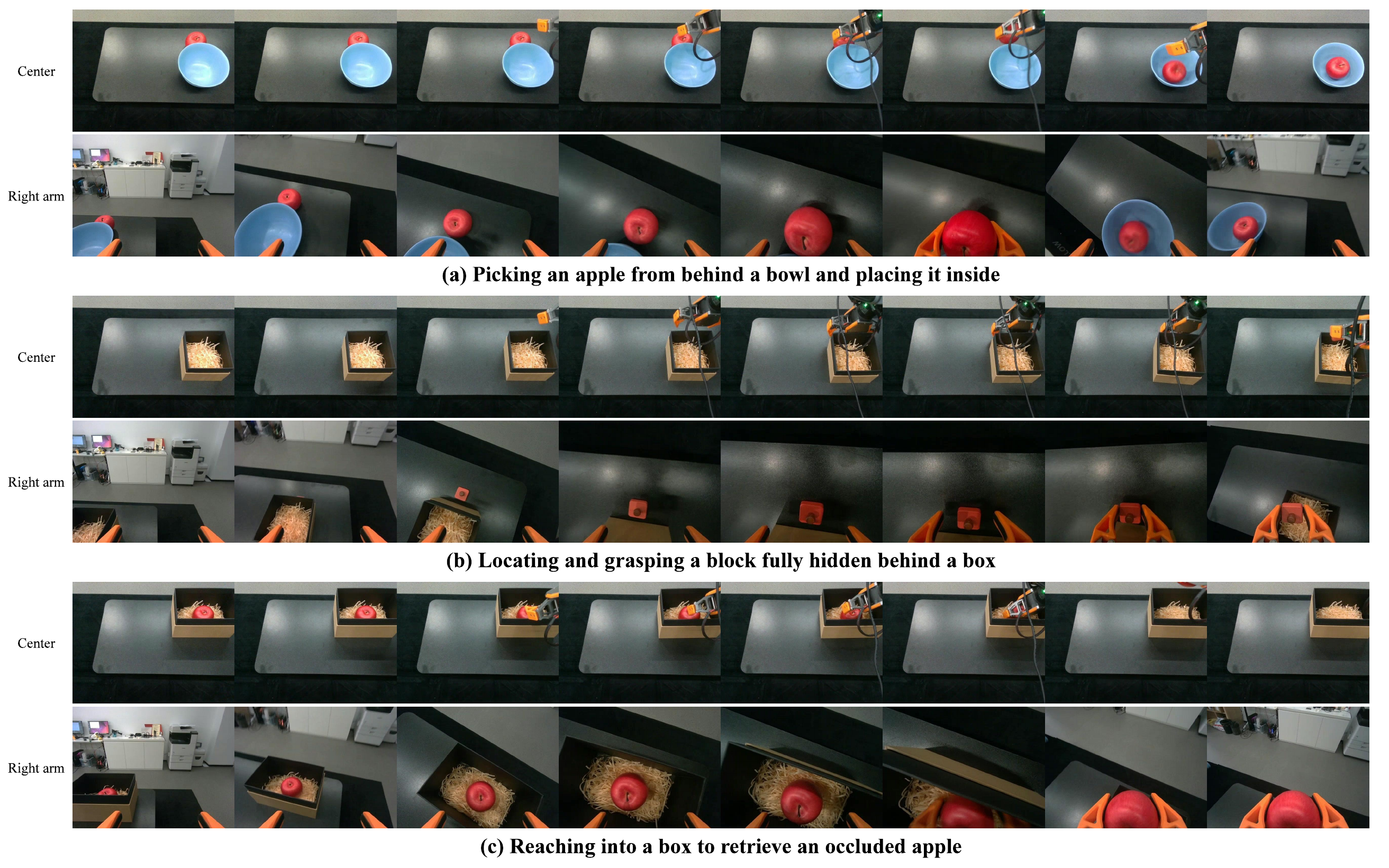}
\caption{\textbf{Physical rollouts with $\pi_{0.5}$ + ActiveAlign.} > Rows show different tasks at varying occlusion levels, with active observation preceding manipulation.}
\label{fig:qual_real}
\end{figure}

\noindent\textbf{Physical result interpretation.} The 60 trials per method in \cref{tab:real_diverse} are a separate evaluation and are not merged with the occlusion aggregate. ActiveAlign improves over both baselines on every task, with 17/20 successes for cup grasping, 15/20 for container retrieval, and 13/20 for observe--open--grasp--place. Their per-task gains over Standard are 40, 45, and 40 pp, but the 20-trial intervals remain wide. We therefore use these results as evidence of broader task coverage rather than a precise estimate of performance over arbitrary physical tasks. Besides, the lower success on the longer task illustrates the limits of an informative initial view over an extended manipulation horizon. \Cref{fig:qual_real} shows the representative rollouts.

\Cref{tab:real} aggregates three apple-and-occluder tasks at each severity. ActiveAlign succeeds in 83/90 trials (92.2\%), compared with 35/90 (38.9\%) for Standard and 17/90 (18.9\%) for Naive Active. Their Wilson intervals are [84.8,96.2]\%, [29.5,49.2]\%, and [12.1,28.2]\%, respectively. Relative to Standard, ActiveAlign gains 43.3, 56.7, and 60.0 pp under partial, medium, and severe occlusion.

\FloatBarrier
\section{LIMITATIONS}
\label{sec:limitations}

ActiveAlign is currently limited by its offline data-construction assumptions. Constructing aligned continuations requires a restorable simulator or a reconstructable physical workcell, as well as demonstration prefixes that remain replayable from the endpoints generated by the viewpoint module. Because training trajectories are retained only after successful alignment and replay, the adapted dataset represents a feasibility-conditioned subset of the deployment endpoint distribution. Difficult endpoints may therefore be underrepresented even though they can still occur at deployment. RLBench evidence is perception-only. The current system performs a single viewpoint-selection phase before manipulation. It does not reselect viewpoints in response to moving occluders. Extending observation-aligned adaptation to non-resettable environments, continuous viewpoint selection, and broader physical deployments remains an important direction.

\section{CONCLUSION}
ActiveAlign connects the decision of where to observe with the learned behavior needed to act from that observation. By constructing successful continuations from active-view endpoints, it enables existing VLA architectures to benefit from a pre-manipulation sensing phase. LIBERO and physical results show consistent gains, while matched-state and data-matched controls support benefits beyond prefix shortening and generic viewpoint coverage. The resulting lesson is that viewpoint selection and manipulation supervision should be designed together, even when the two policies are trained separately. Extending this construction beyond replayable prefixes and interleaving observation selection with manipulation are promising next steps.
\bibliographystyle{IEEEtran}
\bibliography{references}

@article{black2024pi0,
  title={$\pi_{0}$: A Vision-Language-Action Flow Model for General Robot Control},
  author={Black, Kevin and Brown, Noah and Driess, Danny and Esmail, Adnan and Equi, Michael and Finn, Chelsea and Fusai, Niccolo and Groom, Lachy and Hausman, Karol and Ichter, Brian and others},
  journal={arXiv preprint arXiv:2410.24164},
  year={2024}
}

@article{kim2024openvla,
  title={{OpenVLA}: An open-source vision-language-action model},
  author={Kim, Moo Jin and Pertsch, Karl and Karamcheti, Siddharth and Xiao, Ted and Balakrishna, Ashwin and Nair, Suraj and Rafailov, Rafael and Foster, Ethan and Lam, Grace and Sanketi, Pannag and others},
  journal={arXiv preprint arXiv:2406.09246},
  year={2024}
}

@inproceedings{brohan2023rt2,
  title={{RT-2}: Vision-language-action models transfer web knowledge to robotic control},
  author={Zitkovich, Brianna and Yu, Tianhe and Xu, Sichun and Xu, Peng and Xiao, Ted and Xia, Fei and Wu, Jialin and Wohlhart, Paul and Welker, Stefan and Wahid, Ayzaan and others},
  booktitle={Conference on Robot Learning},
  pages={2165--2183},
  year={2023},
  organization={PMLR}
}

@article{brohan2022rt1,
  title={{RT-1}: Robotics transformer for real-world control at scale},
  author={Brohan, Anthony and Brown, Noah and Carbajal, Justice and Chebotar, Yevgen and Dabis, Joseph and Finn, Chelsea and Gopalakrishnan, Keerthana and Hausman, Karol and Herzog, Alex and Hsu, Jasmine and others},
  journal={arXiv preprint arXiv:2212.06817},
  year={2022}
}

@article{team2024octo,
  title={{Octo}: An open-source generalist robot policy},
  author={Team, Octo Model and Ghosh, Dibya and Walke, Homer and Pertsch, Karl and Black, Kevin and Mees, Oier and Dasari, Sudeep and Hejna, Joey and Kreiman, Tobias and Xu, Charles and others},
  journal={arXiv preprint arXiv:2405.12213},
  year={2024}
}

@article{intelligence2025pi05,
  title={$\pi_{0.5}$: a Vision-Language-Action Model with Open-World Generalization},
  author={Intelligence, Physical and Black, Kevin and Brown, Noah and Darpinian, James and Dhabalia, Karan and Driess, Danny and Esmail, Adnan and Equi, Michael and Finn, Chelsea and Fusai, Niccolo and others},
  journal={arXiv preprint arXiv:2504.16054},
  year={2025}
}

@article{pertsch2025fast,
  title={{FAST}: Efficient action tokenization for vision-language-action models},
  author={Pertsch, Karl and Stachowicz, Kyle and Ichter, Brian and Driess, Danny and Nair, Suraj and Vuong, Quan and Mees, Oier and Finn, Chelsea and Levine, Sergey},
  journal={arXiv preprint arXiv:2501.09747},
  year={2025}
}

@article{chi2023diffusion,
  title={Diffusion policy: Visuomotor policy learning via action diffusion},
  author={Chi, Cheng and Xu, Zhenjia and Feng, Siyuan and Cousineau, Eric and Du, Yilun and Burchfiel, Benjamin and Tedrake, Russ and Song, Shuran},
  journal={The International Journal of Robotics Research},
  volume={44},
  number={10-11},
  pages={1684--1704},
  year={2025},
  publisher={Sage Publications Sage UK: London, England}
}

@article{ze20243d,
  title={{3D} diffusion policy: Generalizable visuomotor policy learning via simple {3D} representations},
  author={Ze, Yanjie and Zhang, Gu and Zhang, Kangning and Hu, Chenyuan and Wang, Muhan and Xu, Huazhe},
  journal={arXiv preprint arXiv:2403.03954},
  year={2024}
}

@article{bajcsy1988active,
  title={Active perception},
  author={Bajcsy, Ruzena},
  journal={Proceedings of the IEEE},
  volume={76},
  number={8},
  pages={966--1005},
  year={1988},
  publisher={IEEE}
}

@article{chen2011active,
  title={Active vision in robotic systems: A survey of recent developments},
  author={Chen, Shengyong and Li, Youfu and Kwok, Ngai Ming},
  journal={The International Journal of Robotics Research},
  volume={30},
  number={11},
  pages={1343--1377},
  year={2011},
  publisher={Sage Publications Sage UK: London, England}
}

@inproceedings{morrison2019multi,
  title={Multi-view picking: Next-best-view reaching for improved grasping in clutter},
  author={Morrison, Douglas and Corke, Peter and Leitner, J{\"u}rgen},
  booktitle={2019 International Conference on Robotics and Automation (ICRA)},
  pages={8762--8768},
  year={2019},
  organization={IEEE}
}

@inproceedings{goyal2023rvt,
  title={{RVT}: Robotic view transformer for {3D} object manipulation},
  author={Goyal, Ankit and Xu, Jie and Guo, Yijie and Blukis, Valts and Chao, Yu-Wei and Fox, Dieter},
  booktitle={Conference on Robot Learning},
  pages={694--710},
  year={2023},
  organization={PMLR}
}

@article{goyal2024rvt2,
  title={{RVT-2}: Learning precise manipulation from few demonstrations},
  author={Goyal, Ankit and Blukis, Valts and Xu, Jie and Guo, Yijie and Chao, Yu-Wei and Fox, Dieter},
  journal={arXiv preprint arXiv:2406.08545},
  year={2024}
}

@article{wang2024sam2act,
  title={{SAM2Act}: Integrating visual foundation model with a memory architecture for robotic manipulation},
  author={Fang, Haoquan and Grotz, Markus and Pumacay, Wilbert and Wang, Yi Ru and Fox, Dieter and Krishna, Ranjay and Duan, Jiafei},
  journal={arXiv preprint arXiv:2501.18564},
  year={2025}
}

@article{yu2023visioninaction,
  title={Vision in action: Learning active perception from human demonstrations},
  author={Xiong, Haoyu and Xu, Xiaomeng and Wu, Jimmy and Hou, Yifan and Bohg, Jeannette and Song, Shuran},
  journal={arXiv preprint arXiv:2506.15666},
  year={2025}
}

@inproceedings{chen2025visoGrasp,
  title={{VISO-Grasp}: vision-language informed spatial object-centric 6-DoF active view planning and grasping in clutter and invisibility},
  author={Shi, Yitian and Wen, Di and Chen, Guanqi and Welte, Edgar and Liu, Sheng and Peng, Kunyu and Stiefelhagen, Rainer and Rayyes, Rania},
  booktitle={2025 IEEE/RSJ International Conference on Intelligent Robots and Systems (IROS)},
  pages={14931--14938},
  year={2025},
  organization={IEEE}
}

@inproceedings{viewactive2024,
  title={{ViewActive}: Active viewpoint optimization from a single image},
  author={Wu, Jiayi and Lin, Xiaomin and He, Botao and Ferm{\"u}ller, Cornelia and Aloimonos, Yiannis},
  booktitle={2025 IEEE/RSJ International Conference on Intelligent Robots and Systems (IROS)},
  pages={11812--11818},
  year={2025},
  organization={IEEE}
}

@inproceedings{zhang2025gendop,
  title={{GenDoP}: Auto-regressive camera trajectory generation as a director of photography},
  author={Zhang, Mengchen and Wu, Tong and Tan, Jing and Liu, Ziwei and Wetzstein, Gordon and Lin, Dahua},
  booktitle={Proceedings of the IEEE/CVF International Conference on Computer Vision},
  pages={18229--18239},
  year={2025}
}

@article{liu2024libero,
  title={{LIBERO}: Benchmarking knowledge transfer for lifelong robot learning},
  author={Liu, Bo and Zhu, Yifeng and Gao, Chongkai and Feng, Yihao and Liu, Qiang and Zhu, Yuke and Stone, Peter},
  journal={Advances in Neural Information Processing Systems},
  volume={36},
  pages={44776--44791},
  year={2023}
}

@article{li2026liberoocc,
  title={{LIBERO-Occ}: Evaluating and Improving Vision-Language-Action Models under Scene-Induced Occlusion via Viewpoint Imagination},
  author={Li, Taishan and Zhang, Jiwen and Wang, Siyuan and Wei, Zhongyu and Huang, Xuanjing},
  journal={arXiv preprint arXiv:2606.10862},
  year={2026}
}

@article{james2020rlbench,
  title={{RLBench}: The robot learning benchmark \& learning environment},
  author={James, Stephen and Ma, Zicong and Arrojo, David Rovick and Davison, Andrew J},
  journal={IEEE Robotics and Automation Letters},
  volume={5},
  number={2},
  pages={3019--3026},
  year={2020},
  publisher={IEEE}
}

@article{zhao2023aloha,
  title={Learning fine-grained bimanual manipulation with low-cost hardware},
  author={Zhao, Tony Z and Kumar, Vikash and Levine, Sergey and Finn, Chelsea},
  journal={arXiv preprint arXiv:2304.13705},
  year={2023}
}

@inproceedings{ross2011dagger,
  title={A reduction of imitation learning and structured prediction to no-regret online learning},
  author={Ross, St{\'e}phane and Gordon, Geoffrey and Bagnell, Drew},
  booktitle={Proceedings of the fourteenth international conference on artificial intelligence and statistics},
  pages={627--635},
  year={2011},
  organization={JMLR Workshop and Conference Proceedings}
}

@article{laskin2020reinforcement,
  title={Reinforcement learning with augmented data},
  author={Laskin, Misha and Lee, Kimin and Stooke, Adam and Pinto, Lerrel and Abbeel, Pieter and Srinivas, Aravind},
  journal={Advances in neural information processing systems},
  volume={33},
  pages={19884--19895},
  year={2020}
}

@inproceedings{hansen2021generalization,
  title={Generalization in reinforcement learning by soft data augmentation},
  author={Hansen, Nicklas and Wang, Xiaolong},
  booktitle={2021 IEEE International Conference on Robotics and Automation (ICRA)},
  pages={13611--13617},
  year={2021},
  organization={IEEE}
}

@inproceedings{tobin2017domain,
  title={Domain randomization for transferring deep neural networks from simulation to the real world},
  author={Tobin, Josh and Fong, Rachel and Ray, Alex and Schneider, Jonas and Zaremba, Wojciech and Abbeel, Pieter},
  booktitle={2017 IEEE/RSJ international conference on intelligent robots and systems (IROS)},
  pages={23--30},
  year={2017},
  organization={IEEE}
}

@article{yu2023scaling,
  title={Scaling robot learning with semantically imagined experience},
  author={Yu, Tianhe and Xiao, Ted and Stone, Austin and Tompson, Jonathan and Brohan, Anthony and Wang, Su and Singh, Jaspiar and Tan, Clayton and Peralta, Jodilyn and Ichter, Brian and others},
  journal={arXiv preprint arXiv:2302.11550},
  year={2023}
}

@article{chen2023genaug,
  title={{GenAug}: Retargeting behaviors to unseen situations via generative augmentation},
  author={Chen, Zoey and Kiami, Sho and Gupta, Abhishek and Kumar, Vikash},
  journal={arXiv preprint arXiv:2302.06671},
  year={2023}
}

@article{black2023demogen,
  title={Zero-shot robotic manipulation with pretrained image-editing diffusion models},
  author={Black, Kevin and Nakamoto, Mitsuhiko and Atreya, Pranav and Walke, Homer and Finn, Chelsea and Kumar, Aviral and Levine, Sergey},
  journal={arXiv preprint arXiv:2310.10639},
  year={2023}
}

@inproceedings{connolly1985determination,
  title={The determination of next best views},
  author={Connolly, Cl},
  booktitle={Proceedings. 1985 IEEE international conference on robotics and automation},
  volume={2},
  pages={432--435},
  year={1985},
  organization={IEEE}
}

@inproceedings{doumanoglou2016recovering,
  title={Recovering 6D object pose and predicting next-best-view in the crowd},
  author={Doumanoglou, Andreas and Kouskouridas, Rigas and Malassiotis, Sotiris and Kim, Tae-Kyun},
  booktitle={Proceedings of the IEEE conference on computer vision and pattern recognition},
  pages={3583--3592},
  year={2016}
}

@article{xu2026hommi,
  title={{HoMMI}: Learning Whole-Body Mobile Manipulation from Human Demonstrations},
  author={Xu, Xiaomeng and Park, Jisang and Zhang, Han and Cousineau, Eric and Bhat, Aditya and Barreiros, Jose and Wang, Dian and Bohg, Jeannette and Song, Shuran},
  journal={arXiv preprint arXiv:2603.03243},
  year={2026}
}

@article{zou2026activeglasses,
  title={{ActiveGlasses}: Learning Manipulation with Active Vision from Ego-centric Human Demonstration},
  author={Zou, Yanwen and Shi, Chenyang and Yu, Wenye and Xue, Han and Lv, Jun and Pan, Ye and Wen, Chuan and Lu, Cewu},
  journal={arXiv preprint arXiv:2604.08534},
  year={2026}
}

@article{bjorck2025groot,
  title={{GR00T N1}: An Open Foundation Model for Generalist Humanoid Robots},
  author={{NVIDIA} and Bjorck, Johan and Casta{\~n}eda, Fernando and Cherniadev, Nikita and Da, Xingye and Ding, Runyu and Fan, Linxi and Fang, Yu and Fox, Dieter and others},
  journal={arXiv preprint arXiv:2503.14734},
  year={2025}
}

@inproceedings{mandlekar2023mimicgen,
  title={{MimicGen}: A Data Generation System for Scalable Robot Learning using Human Demonstrations},
  author={Mandlekar, Ajay and Nasiriany, Soroush and Wen, Bowen and Akinola, Iretiayo and Narang, Yashraj and Fan, Linxi and Zhu, Yuke and Fox, Dieter},
  booktitle={Proceedings of the 7th Conference on Robot Learning},
  volume={229},
  series={Proceedings of Machine Learning Research},
  pages={1820--1864},
  publisher={PMLR},
  year={2023}
}

@article{xue2025demogen,
  title={{DemoGen}: Synthetic Demonstration Generation for Data-Efficient Visuomotor Policy Learning},
  author={Xue, Zhengrong and Deng, Shuying and Chen, Zhenyang and Wang, Yixuan and Yuan, Zhecheng and Xu, Huazhe},
  journal={arXiv preprint arXiv:2502.16932},
  year={2025}
}

@article{liu2025reasongrounder,
  title={{ReasonGrounder}: {LVLM}-Guided Hierarchical Feature Splatting for Open-Vocabulary {3D} Visual Grounding and Reasoning},
  author={Liu, Zhenyang and Wang, Yikai and Zheng, Sixiao and Pan, Tongying and Liang, Longfei and Fu, Yanwei and Xue, Xiangyang},
  journal={arXiv preprint arXiv:2503.23297},
  year={2025}
}

@article{liu2026activevla,
  title={{ActiveVLA}: Injecting Active Perception into Vision-Language-Action Models for Precise {3D} Robotic Manipulation},
  author={Liu, Zhenyang and Gu, Yongchong and Wang, Yikai and Xue, Xiangyang and Fu, Yanwei},
  journal={arXiv preprint arXiv:2601.08325},
  year={2026}
}

@article{liu2025trivla,
  title={{TriVLA}: A Triple-System-Based Unified Vision-Language-Action Model with Episodic World Modeling for General Robot Control},
  author={Liu, Zhenyang and Gu, Yongchong and Zheng, Sixiao and Fu, Yanwei and Xue, Xiangyang and Jiang, Yu-Gang},
  journal={arXiv preprint arXiv:2507.01424},
  year={2025}
}

@inproceedings{liu2025neural,
  title={A Neural Representation Framework with {LLM}-Driven Spatial Reasoning for Open-Vocabulary {3D} Visual Grounding},
  author={Liu, Zhenyang and Zheng, Sixiao and Chen, Siyu and Zhao, Cairong and Liang, Longfei and Xue, Xiangyang and Fu, Yanwei},
  booktitle={Proceedings of the 33rd ACM International Conference on Multimedia},
  pages={1042--1051},
  year={2025}
}

@inproceedings{liu2025spatial,
  title={Spatial-Temporal Aware Visuomotor Diffusion Policy Learning},
  author={Liu, Zhenyang and Wang, Yikai and Wang, Kuanning and Liang, Longfei and Xue, Xiangyang and Fu, Yanwei},
  booktitle={Proceedings of the IEEE/CVF International Conference on Computer Vision (ICCV)},
  pages={7122--7131},
  year={2025}
}

@article{liu2026omnicam,
  title={{OmniCam}: Omni-Camera Trajectory Generation via Geometry-Grounded Pose Token Learning},
  author={Liu, Zhenyang and Cao, Chenjie and Zhang, Yisu and Zuo, Xuhui and Xue, Xiangyang and Fu, Yanwei and Wang, Tengfei and Guo, Chunchao},
  journal={arXiv preprint arXiv:2610.09513},
  year={2026}
}

 \end{document}